# Image Measurement Method for Automatic Insertion of Forks into Inclined Pallet

Nobuyuki Kita and Takuro Kato

*Abstract*— In order to insert a fork into a hole of a pallet by a forklift located in front of a pallet, it is necessary to control the height position, reach position, and tilt angle of the fork to match the position and orientation of the hole of the pallet. In order to make AGF (Autonomous Guided Forklift) do this automatically, we propose an image measurement method to measure the pitch inclination of the pallet in the camera coordinate system from an image obtained by using a wide-angle camera. In addition, we propose an image measurement method to easily acquire the calibration information between the camera coordinate system and the fork coordinate system necessary to apply the measurements in the camera coordinate system to the fork control. In the experiment space, a wide-angle camera was fixed at the backrest of a reach type forklift. The wide-angle images taken by placing a pallet in front of the camera were processed. As a result of evaluating the error by comparing the image measurement value with the hand measurement value when changing the pitch inclination angle of the pallet, the relative height of the pallet and the fork, and whether the pallet is loaded or not, it was confirmed that the error was within the allowable range for safely inserting the fork.

## I. INTRODUCTION

In the logistics industry, the amount of work is increasing explosively, while the number of workers is decreasing. Therefore, automation of work is desperately desired.

Efforts are being made to automate the movement of goods from warehouses and the loading and unloading of goods received by trucks. We have conducted research to detect pallets from camera images and estimate their position and orientation for automatic article transfer by AGF. In [1][2], we have developed a method to detect pallets stored on shelves or floor in warehouses from wide-angle images taken from several meters away and estimate the position and orientation of pallets in the camera coordinate system. By using this method, it is possible to automatically drive the AGF in a position and orientation directly facing the pallet to be handled. In [1][2], however, assuming that the shelf or floor surface is horizontal, the orientation of the pallet is estimated only by the angle around the vertical axis (yaw angle).

On the other hand, when a pallet is loaded on a truck, the attitude of the pallet is not horizontal. Therefore, for the AGF located in front of the pallet to insert the fork into the hole of the pallet, it is necessary to control the height position, reach position, and tilt angle of the fork in accordance with the position and attitude of the hole of the pallet (Fig. 1). To this end, it is necessary to measure the pitch angle of the pallet.

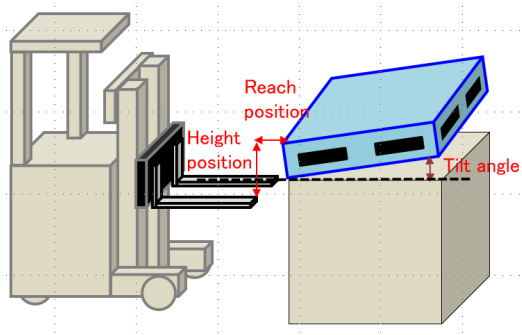

Figure 1. Necessary values for forks to insert the hole of an inclined pallet.

In this paper, an image measurement method capable of precisely measuring a pitch inclination of a pallet in a camera coordinate system from a wide-angle image obtained by using a wide-angle camera, whether goods are placed on the pallet or not, is proposed. In addition, an image measurement method capable of easily and precisely acquiring calibration information between a camera coordinate system and a fork coordinate system required for applying a position and an attitude including a pitch inclination of the pallet in the camera coordinate system to fork control is proposed.

In the experiment space, a wide-angle camera was fixed at center of the upper end of the backrest of a reach type forklift. The wide-angle images taken by placing a pallet in front of the camera were processed. The error was evaluated by comparing the image measurement value and the manual measurement value when changing the pitch angle of the pallet, the relative height between the pallet and the fork, and whether the pallet is loaded or not. As a result, it was confirmed that the error was within the tolerance for safely inserting the fork.

The contribution of the proposed method is as follows.

1.  The pitch inclination of the pallet can be measured with high accuracy regardless of whether the pallet is loaded or unloaded.

2.  Easily calibrate camera and fork with high accuracy.

3.  Enable the automatic insertion of forks into a pallet with an inclined pitch.

The related research is outlined in Chapter 2, pitch inclination measurement methods are explained in Chapter 3, and calibration methods are explained in Chapter 4. Measurement errors are evaluated in Chapter 5 by describing laboratory experiments and results. Finally, Chapter 6 summarizes and prospects.

## II. RELATED WORKS

Methods for measuring the position and orientation of a three-dimensional (3D) object of known shape have been studied for a long time and are introduced in many textbooks. This paper gives an overview of the related studies with a focus on palette.

There have been many studies on pallet detection and position/orientation measurement for the purpose of automation of cargo handling work. They use visual sensors and distance sensors to measure the position/orientation of the pallet in the sensor coordinate system. Many of them assume that the pallet is placed horizontally on a floor or shelf [1]-[13].

Many studies have been conducted to automatically drive a forklift in front of a pallet based on values measured by sensors [3][5][6][8][9]. For this purpose, it is necessary to convert the measured values in the sensor coordinate system into a fork coordinate system or the like. However, few studies have clearly described the coordinate conversion method [4][6][7][10]. There is no research which proposes an original calibration method.

There are only a few research of measuring the inclination of the pallet in the pitch direction with respect to the fork. The author only knows [14]. In this method, the point cloud of the front of the load mounted on the pallet is acquired by a depth sensor, a plane is applied to measure the inclination of the load, and the inclination of the pallet is obtained from the relationship between the front of the load and the pallet. However, this method cannot be applied when no load is mounted on the pallet

There is no research which measured the inclination of the pallet against the fork by the image like this proposal.

## III. IMAGE MEASUREMENT METHOD OF PALLET PITCH INCLINATION

One of the authors has developed a method for estimating the camera inclination during shooting from a panoramic image (equirectangular image) and correcting the inclination [15]. The method detects a group of vertical lines around the image and estimates the camera inclination assuming that most of them are in the vertical direction. According to this method, if a group of two or more parallel lines can be specified on the image, the direction vector of the group of lines in the camera coordinate system can be estimated.

The image shown in Fig. 2(a) is obtained by the wide-angle camera mounted on the backrest (Fig. 2 (b)) when the AGF reaches the position and attitude directly facing the pallet. In this image, the pallet to be handled or the cargo loaded on the pallet appears.

The basic method is to obtain a 3D direction vector of a group of straight lines in a direction close to the upward direction of a panoramic image. In the case of a panoramic image obtained by converting a wide-angle image when carton boxes are loaded on a pallet, the vertical edges of the left and right ends of the load (in the red rectangles in Fig. 3 (a)) are to be analyzed. However, in the case of an unloaded pallet, the vertical left and right edges of the front surface of the pallet become detection candidates (in the yellow rectangles in Fig. 3 (a)), but these lines become very short on the image and are greatly affected by noise. On the other hand, when a panoramic image is created from the wide-angle image with the optical axis direction as the center axis, the left and right edges of the upper surface of the pallet become a group of straight lines in a direction close to the upward direction of the panoramic image and are obtained as lines longer than that (in the red rectangle in Fig. 3 (b)), so that the measurement accuracy is improved.

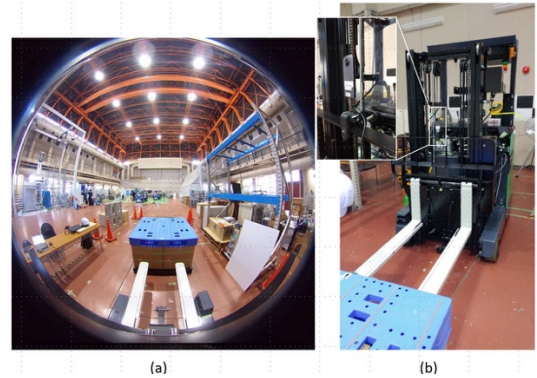

Figure 2. Wide-angle image and a wide-angle camera.

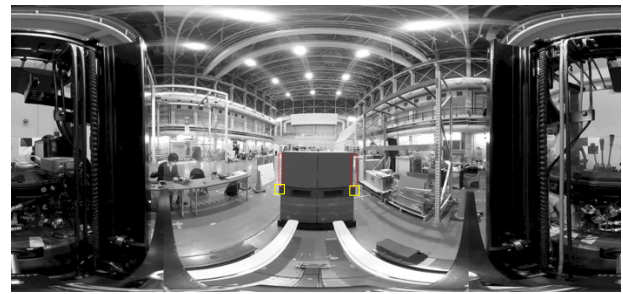

(a)

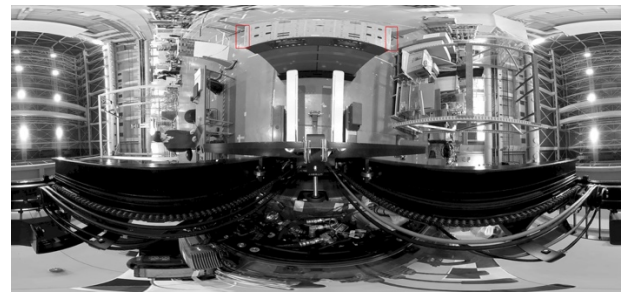

(b)

Figure 3. Panoramic images and the predicted areas for the 3D straight lines.

Fig. 4 shows the definition of the coordinate system. The camera coordinate system is a right-hand system in which the optical center is the origin, the optical axis direction is the X-axis, and the upward direction is the Z-axis. The fork coordinate system is a right-hand system in which the midpoint of the line segment connecting the tips of the left and right forks is the origin, the direction of the line segment is the Y-axis, and the normal direction of the upper surface of the fork is the Z-axis.

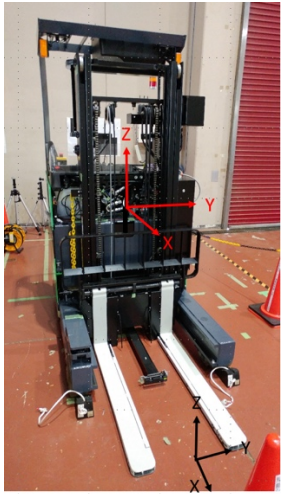

Figure 4. The definitions of the camera and fork coordinate systems.

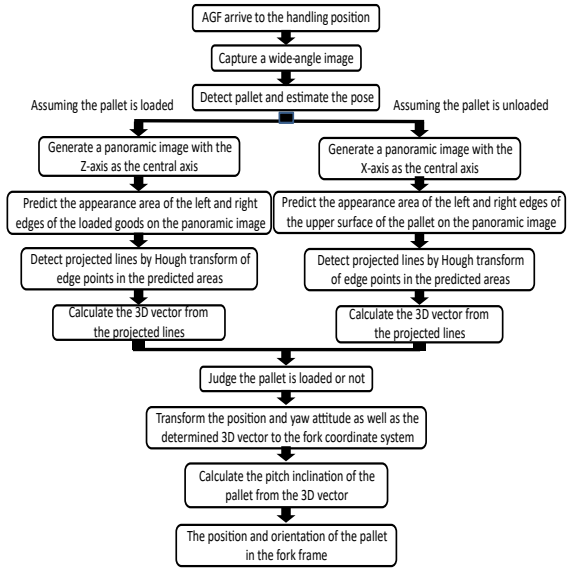

Figure 5. The flowchart of the image measurement method of pallet pitch inclination.

Fig. 5 shows a flow chart of the process. After moving to a position directly opposite to the pallet to be handled, a wide-angle image is acquired, and the pallet is detected from the image and the 3D position and yaw angle are estimated by the method [1][2].

Assuming that the pallet is loaded with goods, the wide-angle image is converted into a panoramic image with the Z-axis direction as the central axis. Since the position and orientation of the pallet in the camera coordinate system are known, the appearance area of the left and right edges of the loaded goods on the panoramic image is predicted (in red rectangle in Fig. 3 (a)). Edge points in the predicted area are Hough transformed by the method of [15] to detect three dimensional straight lines. The direction vector is obtained from the group of parallel 3D straight lines detected in the left and right areas.

Assuming that the pallet is not loaded with goods, the wide-angle image is converted into a panoramic image with the X-axis direction as the center axis. Since the position and orientation of the pallet in the camera coordinate system are known, appearance areas of left and right edges of the upper surface of the pallet on the panoramic image are predicted (red rectangle in Fig. 3 (b)), edge points in the predicted areas are Hough transformed by the method of [15] to detect three dimensional straight lines. The direction vector is obtained from the group of parallel 3D straight lines detected in the left and right areas.

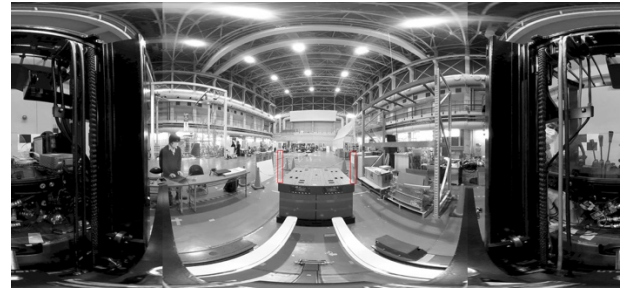

(a)

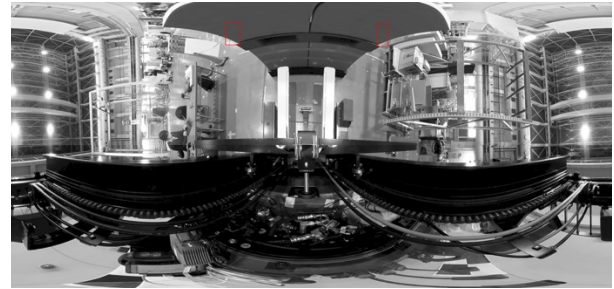

(b)

Figure 6. Panoramic images and the predicted areas for the 3D straight lines.

The score obtained when edge points in the prediction region are subjected to Hough transform corresponds to the length of the 2D projection line. Since the vertical length of the prediction region is an ideal length, it can be determined whether a line exists in the prediction region by the ratio of the score to the vertical length.

When goods are not loaded and processing is performed assuming that goods are loaded, as shown in Fig. 6 (a), a sufficient edge does not exist in the prediction area and this ratio becomes small, so that it can be determined that goods are not loaded. When goods are loaded and processing is performed assuming that goods are not loaded, as shown in Fig. 6 (b), the right and left edges of the upper surface of the pallet in the prediction area are hidden by goods, and the ratio of the score becomes small, so that it can be determined that goods are loaded. In many cases, one or the other will get a valid result, so we adopt that one.

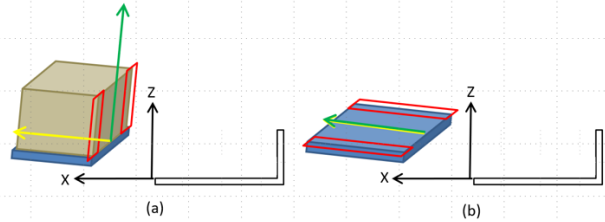

Figure 7. How to calculate the pitch inclination of the pallet from the 3D vector.

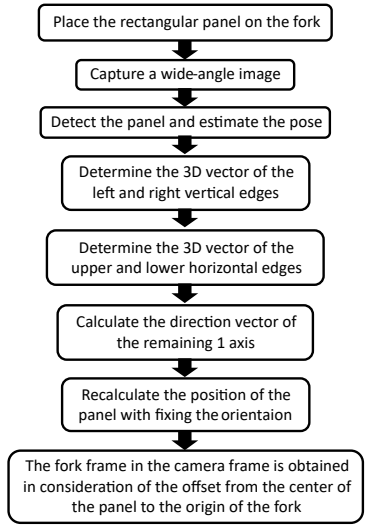

Figure 8. Flow of the calibration process.

However, when many vertical straight lines exist in the background, a straight line may be detected in the prediction area in a panoramic image assuming that goods are loaded, even if goods are not loaded. Therefore, if valid results are obtained in both cases, the result obtained by assuming goods are not loaded is adopted.

Using the calibration information described in the next section, the position and yaw attitude of the pallet as well as the determined direction vector are transformed from the camera coordinate system to the fork coordinate system (green arrow in Fig. 7). The pitch inclination of the pallet is calculated from the 3D vector obtained by rotating the direction vector by 90 degrees (yellow arrow in Fig. 7 (a)) when the pallet is loaded, and from the determined direction vector itself (yellow arrow in Fig. 7 (b)) when the pallet is unloaded.

## IV. CAMERA AND FORK COORDINATE SYSTEM CALIBRATION

In this proposal, it is assumed that the camera is fixed to the backrest of a reach type forklift. Since the backrest is integrated with the fork, the position and orientation relationship between the camera and the fork does not change before and after the movement of the fork. Therefore, it is only necessary to calibrate once before the start of cargo handling work. However, the relationship between the camera and the fork, especially the orientation relationship, was difficult to obtain with high precision because of the small size of the camera housing. We propose a simple calibration method using wide-angle images.

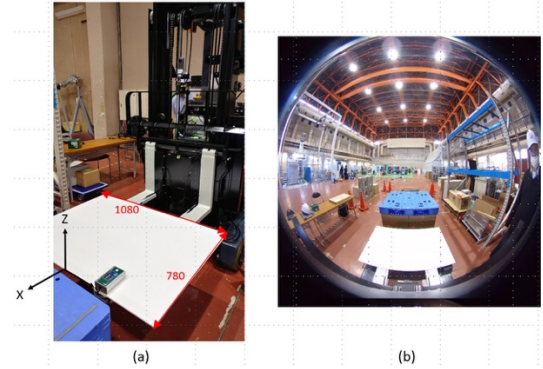

Figure 9. Panel and the wide-angle image for the calibration.

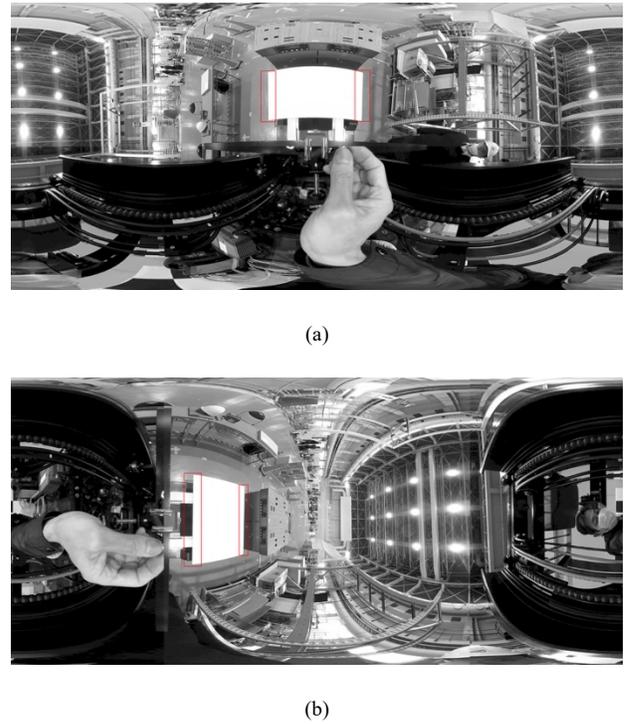

Figure 10. Panoramic images and the predicted areas for the panel edges.

### A. Calibration Method

Fig. 8 shows a flow of the calibration process. Place a rectangular panel of known dimensions on the fork so that the center of its front lateral edge coincides with the origin of the fork coordinate system and its vertical and horizontal edges are parallel to the X and Y axes of the fork coordinate system (Fig. 9 (a)). Fig. 9 (b) shows wide-angle image taken.

The panel is detected and the position and orientation of the panel in the camera coordinate system are estimated by the method of [1][2].

Similarly to the determination of the direction vectors of the left and right edges of the upper surface of the palette, the X direction of the camera is regarded as a pseudo vertical direction to convert the wide-angle image into a panoramic image (Fig. 10 (a)), and the method of [15] is applied to determine the direction vector in the camera coordinate system of the two left and right vertical edges of the panel.

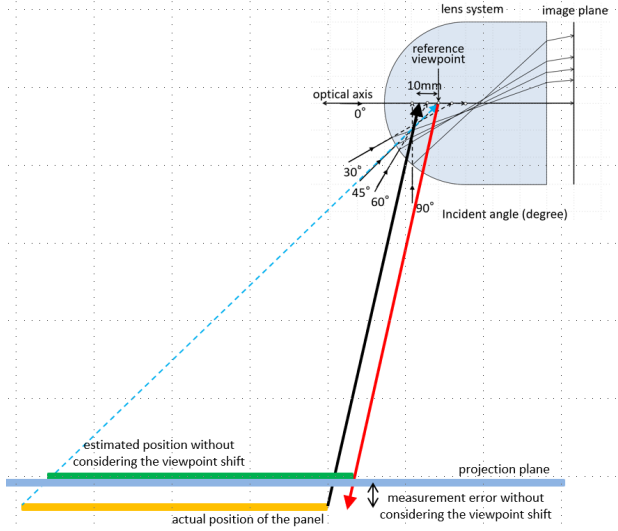

Figure 11. Viewpoint shift and the measurement error.

In order to obtain direction vectors of two horizontal edges on the upper and lower sides of the panel, a wide-angle image is converted into a panorama image (Fig. 10 (b)) by regarding the Y-axis direction of the camera as a pseudo vertical direction, and the method of [15] is applied to obtain direction vectors in the camera coordinate system of two horizontal edges on the upper and lower sides of the panel.

The orientation of the panel in the camera coordinate system is updated by obtaining the direction vector of the remaining 1 axis from the vertical direction vector and the horizontal direction vector of the panel.

In order to improve the position estimation accuracy obtained by the method [2], the posture of the panel in the update, and then the method [2] is applied again to obtain the position of the panel in the camera coordinate system.

Finally, from the position and orientation of the panel in the camera coordinate system obtained, the position and orientation of the fork coordinate system in the camera coordinate system is obtained in consideration of the offset from the center of the panel to the origin of the fork coordinate system.

As described above, simply by placing a rectangular panel of known size on a fork and taking an image, the position and orientation of the fork coordinate system in the camera coordinate system can be obtained with high accuracy by performing program processing.

### B. Error due to the Presence or Absence of Consideration of Viewpoint Shift

[1][2] create a projection image on an arbitrary projection plane from a wide-angle image, compares it with a model of a palette (or panel), and obtains the position and orientation of the palette (or panel) from the position and orientation of the projection plane having the maximum similarity.

As described in [16], the viewpoint of a fish-eye lens for obtaining a wide-angle image moves according to the angle of incident light. Fig. 11 shows the relationship between the camera and the panel at the time of calibration. In the case of a wide-angle camera, if the incident position when the incident angle is 45 degrees (blue dotted line) is taken as the reference viewpoint, a light ray (black) with an incident angle of about 80 degrees is incident at a position offset from the reference viewpoint by about 10 mm in the forward direction. If calibration is performed without taking this viewpoint shift into consideration, the estimation of the height position of the panel is significantly adversely affected. Assume that the orange color in Fig. 11 is the actual position of the panel. The light ray from the lower horizontal edge of the panel is incident at a position offset from the reference viewpoint by about 10 mm as shown by the black arrow. On the other hand, if the viewpoint shift is not taken into consideration, the light ray corresponding to the image of the lower horizontal edge of the panel is regarded as incident from the reference viewpoint as shown by red arrow. Therefore, if the projection image (green) is generated while shifting the projection plane (blue) up and down and without considering the viewpoint shift, as shown in the Fig. 11, the similarity between the projection image and the model at a position 10 mm higher than the actual position becomes the maximum, and an error of 10 mm occurs in the measurement of the height position.

## V. EXPERIMENTS

Wide-angle images acquired by a wide-angle camera (Ricoh ThetaV) fixed at the center of the upper end of the backrest of a reach type forklift in an indoor experiment space were processed by a notebook PC (Thinkpad X1 Extreme).

Like this paper, there is a study [14] in which an experiment to measure the pitch inclination of a pallet is carried out in order to insert a fork into the hole of the pallet. However, it does not carry out a systematic experimental evaluation for different conditions. In the first place, it is impossible to compare the results because the sensors used are different.

Therefore, it was decided to evaluate the experimental results by using as an index whether the fork can be safely inserted into the hole of the pallet.

### A. Evaluation of tolerances for safe insertion of forks

Fig. 12 shows the sizes of forks and target pallets. As shown in Fig. 13, the operation of inserting the fork into the pallet is realized by the following two steps.

1) Based on the estimated pallet position and orientation, the position and orientation of the fork are controlled so that the center position of the tip of the fork becomes the center position of the opening of the pallet and the tilt angle of the fork becomes the same angle as the pitch angle of the pallet.

2) Insert the fork straight along the direction of tilt inclination.

With respect to the estimated position and orientation values, $\Delta X$ is an error in the reach direction, $\Delta Z$ is an error in the height direction, and $\Delta \theta$ is an error in the tilt angle. As shown in 1) in Fig. 13, these errors are the difference between the position and orientation angle of the fork and the pallet at the time when the operation 1) is completed.

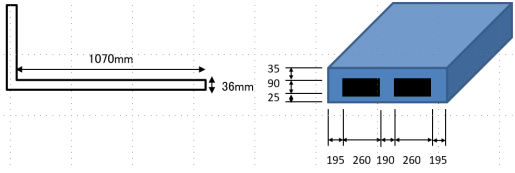

Figure 12. The sizes of forks and target pallets.

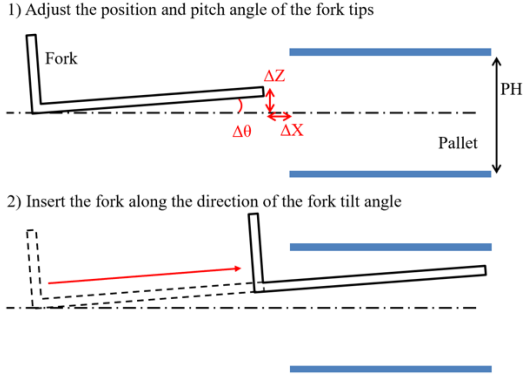

Figure 13. The operation of inserting the fork into the pallet.

It is shown in [2] that the position estimation error of the pallet is 20 [mm] or less. Assuming that $\Delta X = 20$ [mm], the difference in the height of the fork tip at the completion of the insertion operation between the case of $\Delta \theta = 0$ [deg] and the case of $\Delta \theta = 1$ [deg] is calculated as $\Delta X \cdot tan(\Delta \theta \cdot \frac{2\pi}{360}) = 0.3491 [mm]$.

That is, the change in the fork height caused by the error $\Delta X$ in the reach direction is negligibly small. In the following description, the conditions for inserting the fork without contacting the pallet are calculated in consideration of only the error $\Delta Z$ in the height direction and the error $\Delta \theta$ in the tilt angle.

The condition for $\Delta Z$ and $\Delta \theta$ that the fork does not interfere with the pallet during the insertion operation is that the following four inequalities are satisfied.

$$-\frac{PH}{2} < \Delta Z - \left(\frac{FT}{2}\right) \cdot \cos(\Delta \theta)$$

$$\Delta Z + \left(\frac{FT}{2}\right) \cdot \cos(\Delta \theta) < \frac{PH}{2}$$

$$-\frac{PH}{2} < \Delta Z - \left(\frac{FT}{2}\right) \cdot \cos(\Delta \theta) + FL \cdot \sin(\Delta \theta)$$

$$\Delta Z + \left(\frac{FT}{2}\right) \cdot \cos(\Delta Z) + FL \cdot \sin(\Delta \theta) < \frac{PH}{2}$$

$PH$ is the height of the pallet's insertion slot, $FT$ is the thickness of the fork, and $FL$ is the length of the fork. As shown in Fig. 12, $PH$ is $90[mm]$, $FT$ is $36$ [mm], and $FL$ is $1070$ [mm] in our case. Combinations of $\Delta Z$ and $\Delta \theta$ that satisfies such condition is an internal region of a square shown in blue in Fig. 14.

### B. Calibration Experiment

The internal parameters of the wide-angle camera (Ricoh ThetaV) have already been acquired by the OCamCalib [17].

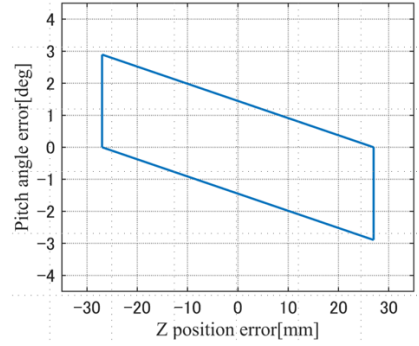

Figure 14. Tolerances for safe insertion of forks.

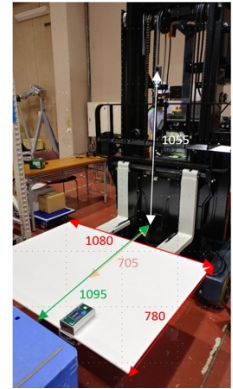

Figure 15. The manually measured positional relationship between the camera and the panel.

Table 1. The obtained calibration information without/with consideration of the viewpoint shift.

|  | panel position in camera frame | | |
|---|---|---|---|
|  | manual measurement | image measurement without shift | with shift |
| Δ X (m) | 0.705 | 0.687 | 0.695 |
| Δ Z (m) | -1.055 | -1.042 | -1.050 |

Fig. 15 shows the manually measured positional relationship between the camera and the panel.

Calibration was performed by the method described in 4.1. As described in 4.2, in order to improve the accuracy of calibration, it is necessary to create a projected image taking into consideration of the viewpoint shift. However, the viewpoint shift curve for the ThetaV has not been obtained. The viewpoint shift curve of another wide-angle camera having similar viewing angle and distortion characteristics was substituted.

Table 1 shows the obtained calibration information.

### C. Pallet Position and Orientation Estimation Experiment in Fork Coordinate System

The pallet was placed in front of the AGF in the experiment space, and 18 wide-angle images were taken while changing the pitch inclination angle of the pallet (three ways), the relative height of the pallet and fork (three ways), and whether the pallet was loaded or not (two ways) (Fig. 16).

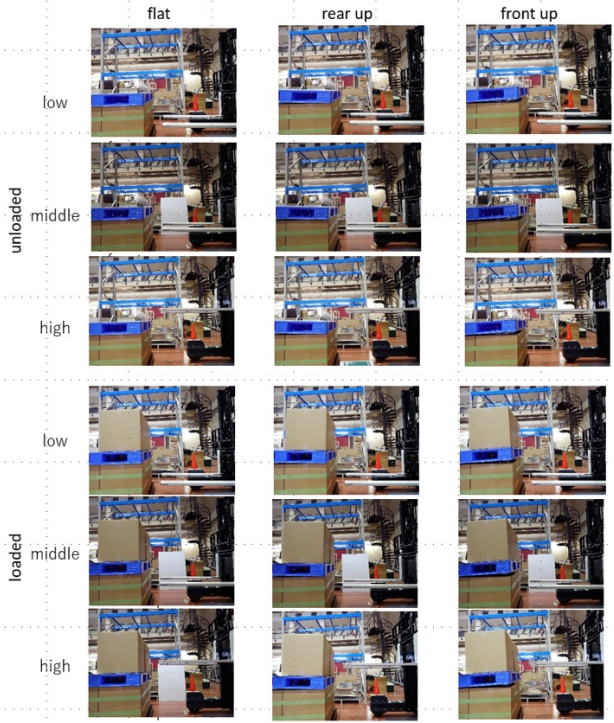

Figure 16. The relation between a pallet and forks for evaluation experiment.

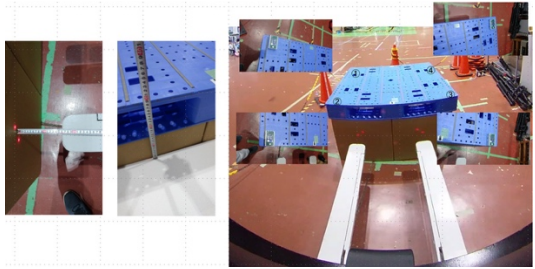

Figure 17. Manual measurement by using a tape measure and a clinometer.

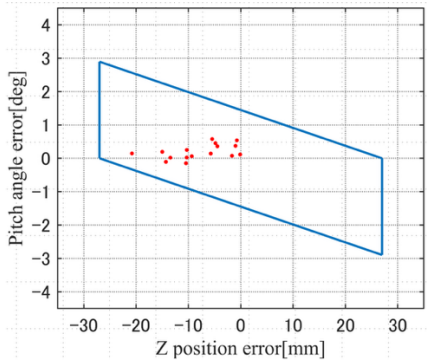

Figure 18. Image measurement results plotted on the tolerance graph.

At the same time, $\Delta X$, $\Delta Z$ and $\Delta \theta$ were manually measured by using a tape measure and a clinometer, as shown in Fig. 17, and those are regarded as the true value.

The results obtained by image processing all and the true values are shown in Table 2. The last two cases failed to detect the palette. For the remaining 16 cases, the set of the error $\Delta Z$ in the height direction and the error $\Delta \theta$ in the tilt angle were plotted with red dots on the tolerance graph (Fig. 18). All 16 cases were confirmed to be within the tolerance for safely inserting the fork.

## VI. Conclusion

This paper proposes an image measurement method to measure the pitch inclination of the pallet in the camera coordinate system by analyzing the wide-angle image. It also proposes an image measurement method to easily obtain calibration information between the camera coordinate system and the fork coordinate system with high accuracy.

In an experiment in which a pallet was placed in front of a forklift in an experimental space, promising results were obtained. In the actual warehouse, another pallet is often placed adjacent to the pallet to be handled. The lighting conditions change variously depending on whether it is outdoors or indoors, or whether it is at a low place or at a high place. In order to apply the proposed method to the real environment, it is necessary to improve the method so that the pitch inclination can be measured robustly even in an environment with complicated changes. For example, the selection of whether to use an edge point in a prediction region for Hough transformation is performed at a fixed threshold value at present, but it can be considered that it is adaptively determined.

We are preparing an experiment to automatically insert a fork into a pallet inclined in the pitch direction by automatically controlling the fork according to the measurement values obtained by the proposed method. The results are expected to be obtained soon.

In the future, the AGF will be automatically controlled to the position and orientation directly facing the target pallet based on the measurement value obtained by the already developed method that detects the pallet from a position more than 4m away and increases the estimation accuracy of the position and orientation of the pallet as it approaches, and the automatic insertion will be performed by the pallet pitch angle measured by the proposed method.

Table 2. The results obtained by image processing, GT and the error.

| with goods? | fork position | pallet pitch | pallet pose in fork frame | | | | | | | | |
|---|---|---|---|---|---|---|---|---|---|---|---|
| | | | $\Delta X$ (mm) | | | $\Delta Z$ (mm) | | | $\Delta \theta$ (degree) | | |
| | | | estimated | GT | error | estimated | GT | error | estimated | GT | error |
| unloaded | low | flat | 221.538 | 201 | 20.538 | 349.269 | 355 | -5.731 | 1.7921 | 1.65 | 0.1421 |
| | | rear up | 225.99 | 201 | 24.99 | 344.695 | 355 | -10.305 | -0.73906 | -0.99 | 0.25094 |
| | | front up | 235.288 | 201 | 34.288 | 390.633 | 400 | -9.367 | 4.25245 | 4.19 | 0.06245 |
| | middle | flat | 228.905 | 203 | 25.905 | 206.206 | 227 | -20.794 | 1.7959 | 1.65 | 0.1459 |
| | | rear up | 216.324 | 203 | 13.324 | 212.008 | 227 | -14.992 | -0.796057 | -0.99 | 0.193943 |
| | | front up | 233.191 | 203 | 30.191 | 258.549 | 272 | -13.451 | 4.21031 | 4.19 | 0.02031 |
| | high | flat | 217.097 | 216 | 1.097 | -274.328 | -264 | -10.328 | 1.67478 | 1.65 | 0.02478 |
| | | rear up | 213.585 | 216 | -2.415 | -278.307 | -264 | -14.307 | -1.09579 | -0.99 | -0.10579 |
| | | front up | 228.186 | 216 | 12.186 | -229.469 | -219 | -10.469 | 4.04312 | 4.19 | -0.14688 |
| loaded | low | flat | 225.998 | 201 | 24.998 | 354.881 | 355 | -0.119 | 1.76497 | 1.65 | 0.11497 |
| | | rear up | 221.68 | 201 | 20.68 | 353.996 | 355 | -1.004 | -0.615755 | -0.99 | 0.374245 |
| | | front up | 230.997 | 201 | 29.997 | 398.318 | 400 | -1.682 | 4.26566 | 4.19 | 0.07566 |
| | middle | flat | 220.582 | 203 | 17.582 | 222.527 | 227 | -4.473 | 2.01217 | 1.65 | 0.36217 |
| | | rear up | 216.537 | 203 | 13.537 | 222.168 | 227 | -4.832 | -0.535763 | -0.99 | 0.454237 |
| | | front up | 224.874 | 203 | 21.874 | 266.538 | 272 | -5.462 | 4.76724 | 4.19 | 0.57724 |
| | high | flat | 210.166 | 216 | -5.834 | -264.746 | -264 | -0.746 | 2.18912 | 1.65 | 0.53912 |
| | | rear up | | 216 | -216 | | -264 | 264 | | -0.99 | 0.99 |
| | | front up | 134.873 | 216 | -81.127 | 8.87414 | -219 | 227.8741 | 3.51914 | 4.19 | -0.67086 |